# Recurrent neural network transducer for Japanese and Chinese offline handwritten text recognition

Trung Tan Ngo[0000-0002-8021-1072], Hung Tuan Nguyen[0000-0003-4751-1302], Nam Tuan Ly[0000-0002-0856-3196], Masaki Nakagawa[0000-0001-7872-156X]

Tokyo University of Agriculture and Technology, Tokyo, Japan
[trungngotan94, ntuanhung, namlytuan]@gmail.com,
nakagawa@cc.tuat.ac.jp

**Abstract.** In this paper, we propose an RNN-Transducer model for recognizing Japanese and Chinese offline handwritten text line images. As far as we know, it is the first approach that adopts the RNN-Transducer model for offline handwritten text recognition. The proposed model consists of three main components: a visual feature encoder that extracts visual features from an input image by CNN and then encodes the visual features by BLSTM; a linguistic context encoder that extracts and encodes linguistic features from the input image by embedded layers and LSTM; and a joint decoder that combines and then decodes the visual features and the linguistic features into the final label sequence by fully connected and softmax layers. The proposed model takes advantage of both visual and linguistic information from the input image. In the experiments, we evaluated the performance of the proposed model on the two datasets: Kuzushiji and SCUT-EPT. Experimental results show that the proposed model achieves state-of-the-art performance on all datasets.

## 1 Introduction

There is an increasing demand for automatic handwriting recognition for different purposes of document processing, educating, and digitizing historical documents in recent years. Over the past decade, numerous fundamental studies have provided large handwriting sample datasets such as handwritten mathematic expressions [1], handwritten answers [2], and historical documents [3, 4]. Moreover, many deep neural network (DNN)-based approaches are proposed to extract visual features for recognizing handwritten text [5–9]. Thanks to the acceleration of computational software and hardware, DNN has become an essential component of any handwriting recognition system.

One of the essential characteristics of handwriting is that there are many groups of categories having similar shapes [10, 11]. It is especially true in languages as Chinese or Japanese, where the number of categories is more than 7000, so that many groups of categories cannot be distinguished alone. Therefore, current handwriting recognition systems need to rely on grammar constraints or dictionary constraints as post-processing to eliminate inaccurate predictions. These post-processing methods have been studied for a long time with several popular methods such as the n-gram model [12], dictionary-driven model [13], finite-state machine-based model [14], recurrent neural network (RNN)-based language model [15].

These linguistic post-processing processes are combined into the DNN in most practical handwriting recognition systems. Generally, the handwriting predictions are combined with the linguistic probabilities in a log-linear way and beam search to get the n-best predictions at the inference process [15]. These methods are usually named as the explicit language models, which need to be pre-trained using large corpora. On the other hand, the language models could also be implicitly embedded into DNN during the training process. In this case, however, the handwriting recognizer might overfit a text corpus if the training dataset is not diverse with many different sentence samples [7].

The limitation of the linguistic post-processing methods is the discreteness of the linguistic post-processing processes from the DNN character/text recognition models. Previous studies have shown that end-to-end unified DNN systems achieve better recognition results than discrete systems [16–18]. This is because every module in an end-to-end system is optimized simultaneously as other DNN modules, while individual systems might not achieve global optimization. Therefore, we propose to directly integrate a linguistic context module into a well-known DNN for the handwriting recognition problem.

The main contribution of this paper is to demonstrate the way and the effectiveness of RNN-Transducer for Chinese and Japanese handwriting recognition, which combines both visual and linguistic features in a single DNN model. As far as we know, this is the first attempt to combine these two different types of features using the RNN transducer model. It might open a new approach for improving handwriting recognition. The rest of the paper is organized as follows: Section 2 introduces the related works while Section 3 describes our methods. Section 4 and 5 present the experiment settings and results. Finally, Section 6 draws our conclusion.

## 2   Related work

Most traditional text recognition methods are based on the segmentation of a page image into text lines and each text line into characters. The latter stage has been studied intensively [19, 20]. However, this segmentation-based approach is error-prone and affects the performance of the whole recognizer. In recent years, based on deep learning techniques, many segmentation-free methods are designed to surpass the problem and proven to be state-of-the-art for many text recognition datasets. These methods are grouped into two main approaches: convolutional recurrent neural network (CRNN)[21] and attention-based sequence-to-sequence.

Graves et al. [21] introduced Connectionist Temporal Classification (CTC) objective function and combined it with convolutional network and recurrent neural network to avoid explicit alignment between input images and their labels. The convolutional network helps extract features from images, while the recurrent network deals with a sequence of labels. This combination achieved high performance in offline handwriting recognition [9].

Besides, the attention-based sequence-to-sequence methods have been successfully adopted in many tasks such as machine translation [22] and speech recognition [23]. These methods are also applied to text recognition tasks and achieve high accuracy [11]. The attention mechanism helps to select the relevant regions of an input image to recognize text sequentially. Furthermore, the methods are also applied for lines by implicit learning the reading order and predict recognized text in any order. However, the attention-based methods require predicting exactly the same as the target result [2]. In addition, these models do not achieve high results in the Chinese and Japanese datasets compared to CRNN and CTC methods [2, 24].

Recently, some new CRNN models combined with the attention mechanism achieved state-of-the-art results in speech recognition [25]. In the Kuzushiji Japanese cursive script dataset, Nam et al. [24] also achieved the best result with the same approach. However, these modifications cannot surpass the result of the CRNN model in the SCUT-EPT Chinese examination dataset [2]. In Chinese and Japanese, a character pattern consists of multiple radicals, and the radical set is much smaller than the character set. Some researchers [26, 27] use the relationship between radicals and characters to enhance recognition. They reported the best results in both CASIA [4] and SCUT-EPT datasets.

The CRNN model has some limitations, such as requiring the output sequence not longer than the input sequence and does not model the interdependencies between the outputs. Therefore, Graves [28] introduced the RNN-Transducer as an extension of CTC to address these mentioned issues.

The RNN-Transducer defines a joint model of both input-output and output-output. Therefore, the RNN-Transducer mixes the visual features from input and context information of predicted result to find later prediction. It can be seen as an integration of a language model into the CRNN model without external data. In addition, the model also defines a distribution over the output sequence without the length constraint. However, there is no prior research about RNN-Transducer in the offline handwritten text recognition field. We will describe this model architecture and its application to Chinese and Japanese handwriting datasets in the next sections.

## 3 RNN-Transducer architecture

Similar to RNN-Transducer in [28], its network architecture in handwriting recognition topic also has three main parts: a visual feature encoder, a linguistic context encoder, and a joint decoder. This structure is depicted in Fig. 1. At first, a convolutional neural network (CNN) extracts local features $x_t$ from an image. Then these features are encoded by a recurrent neural network (RNN) as in the CRNN model. During training, the linguistic encoder extracts context features from previous characters in ground truth text. Meanwhile, it takes input from predicted characters in the inference process. These two feature vectors are aggregated into a joint decoder. Then, this decoder predicts a density over the output sequence distribution. In this experiment, we apply a greedy search for the inference process.

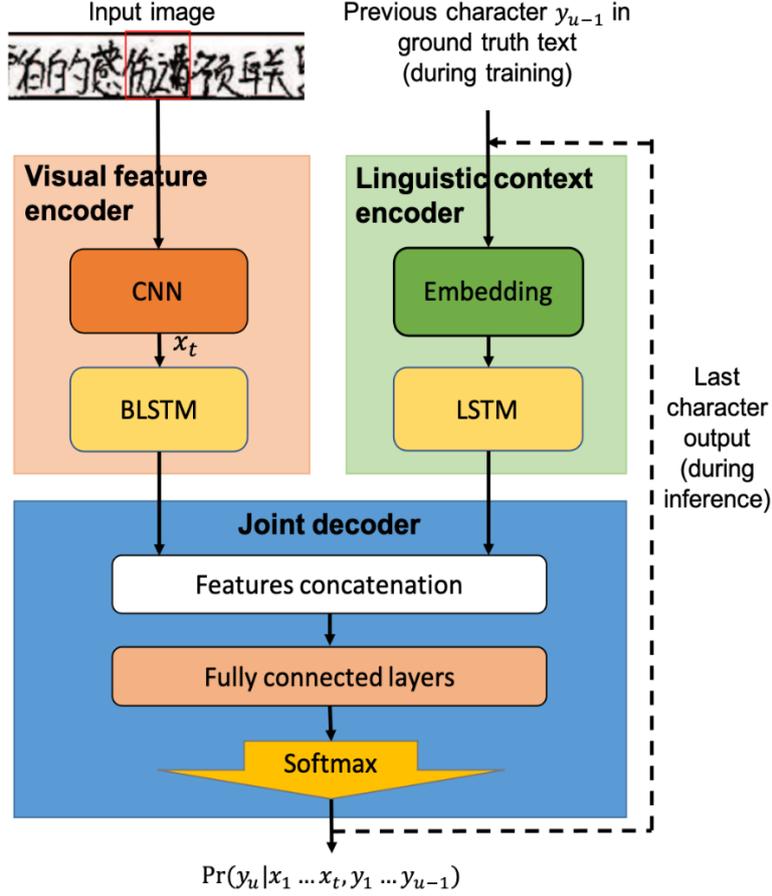

**Fig. 1.** RNN-Transducer architecture.

Let $\boldsymbol{x} = (x_1, x_2, \ldots, x_T), x_t \in \boldsymbol{X}$ be an input feature sequence of length $T$ over the visual feature space $\boldsymbol{X}$. Meanwhile, let $\boldsymbol{y} = (y_1, y_2, \ldots, y_U), y_u \in \boldsymbol{Y}$ be an output sequence of length $U$ over the output space $\boldsymbol{Y}$. In addition, let $K$ is the number of categories, thus output vectors in $\boldsymbol{y}$ have the size $K$.

The extended output space $\overline{\boldsymbol{Y}}$ is defined as $\boldsymbol{Y} \cup \{\emptyset\}$, where $\emptyset$ denotes a blank symbol. The meaning of the blank symbol is "output nothing". Thus, it is used to represent alignments of an output text on an input sample.

Let $\boldsymbol{a} = (a_{1,1}, a_{1,2}, \ldots, a_{T,U}), a_{i,j} \in \overline{\boldsymbol{Y}}$ as alignments of $\boldsymbol{y}$. For example, in case $\boldsymbol{y} = (B, E, E)$ and $\boldsymbol{x} = (x_1, x_2, x_3, x_4)$, some valid alignments are $(\emptyset, \emptyset, B, \emptyset, E, E, \emptyset)$, $(\emptyset, B, E, E, \emptyset, \emptyset, \emptyset)$, $(\emptyset, B, \emptyset, E, \emptyset, E, \emptyset)$, or $(\emptyset, \emptyset, B, E, \emptyset, E, \emptyset)$. Given $\boldsymbol{x}$, the RNN-

Transducer defines the conditional distribution Pr($y|x$) as a sum over all possible alignments $a$ as Eq. (1).

$$Pr(y|x) = \sum_{a \in B^{-1}(y)} Pr(a|x) \qquad (1)$$

where $B$ is a function that removes $\emptyset$ symbols from $a$, which is defined by $B(a) = y$.

### 3.1  Visual feature encoder

In this work, the visual feature encoder is composed of a CNN and an RNN. As mentioned above, the CNN effectively extracts features from handwritten text images [15, 16]. This CNN module is constructed by stacking convolutional, batch norm, and max pool layers. It extracts a feature sequence from a text image by sliding a sub-window through the input image along a defined direction.

This feature sequence is then passed to the recurrent neural network. The RNN allows information of nearby inputs to remain in the network's internal states. Therefore, the RNN can use visual contextual features and support for the following processes. In our proposed visual feature encoder, the BLSTM is used to learn the long-range context in both input directions.

Given a text line image, the CNN extracts a feature sequence $(x_1, x_2, ..., x_T)$. Then the recurrent layers in BLSTM scan forward and backward this sequence and compute a feature sequence $f = (f_1, f_2, ..., f_T)$. This encoder is similar to the CRNN model [9].

### 3.2  Linguistic context encoder

The linguistic context encoder consists of an input embedded layer and an RNN. The inputs are sparsely encoded as one-hot vectors of size $K+1$. At first, the embedded layer maps one-hot variables into dense representations. It reduces the dimensionality of variables and creates meaningful vectors in the transformed space. Then, we use the LSTM for this encoder to process embedded vectors.

This network gets input from a sequence $\mathbf{y} = (y_1, y_2, ..., y_U)$ of length $U$ and outputs a context vector sequence $g = (g_1, g_2, ..., g_U)$. It is similar to a deep learning language model, which extracts linguistic information from the context. These vectors are then aggregated with vectors from the visual feature encoder. Therefore, a dense vector with a smaller size than the character set is more appropriate for the aggregation purpose.

### 3.3  Joint decoder

The joint decoder is constructed by several fully connected layers and a softmax layer. It combines features from the transcription vector $f_t$, where $1 \leq t \leq T$, and the prediction vector $g_u$, where $1 \leq u \leq U$. Then it defines the output distribution over $\overline{Y}$ as Eq. (2).

$$Pr(a_{t,u}|f_t, g_u) = softmax(W^{joint} \tanh(W^{visual} f_t + W^{linguistic} g_u + b)) \quad (2)$$

where $W^{visual}$, $W^{linguistic}$ are the weights of linear projection, in turn, for visual feature vectors and context feature vectors to a joint subspace; $b$ is a bias for this combination; $W^{joint}$ is a mapping to the extended output space $\overline{Y}$. Furthermore, the function $Pr(\boldsymbol{a}|\boldsymbol{x})$ in Eq. (1) is factorized as Eq. (3).

$$Pr(\boldsymbol{a}|\boldsymbol{x}) = \prod_{t=1}^{T} \prod_{u=1}^{U} Pr(a_{t,u}|f_t, g_u) \quad (3)$$

### 3.4 Training and inference process

To train the RNN-Transducer model, the log-loss $L = -\log(Pr(\boldsymbol{y}|\boldsymbol{x}))$ is minimized. However, a naive calculation of $Pr(\boldsymbol{y}|\boldsymbol{x})$ in Eq. (1) by finding all alignments $\boldsymbol{a}$ is intractable. Therefore, based on Eq. (3), Graves et al. used the forward-backward algorithm [28] to calculate both loss and gradients.

In the inference process, we seek the predicted vector $\boldsymbol{a}$ with the highest probability of the input image. However, the linguistic feature vector $g_u$ and hence $Pr(a_{t,u}|f_t, g_u)$ depends on all previous predictions from the model. Therefore, the calculation for all possible sequences $\boldsymbol{a}$ is also intractable. In this paper, we apply the greedy search for a fast inference process through output sequences as shown in Algorithm 1.

**Algorithm 1: Output Sequence Greedy Search**
```
Initalise: Y=[]
g, h = LinguisticEncoder(∅)
for t = 1 to T do
  out = JointDecoder(f[t], g)
  y_pred = argmax(out)
  if y_pred != ∅:
    add y_pred to Y
    g, h = LinguisticEncoder(y_pred, h)
  end if
end for
Return: Y
```

## 4 Datasets

To evaluate the proposed RNN-Transducer model, we conducted experiments on the Kuzushiji_v1 Japanese text line and SCUT-EPT Chinese examination paper text line datasets. Our model is evaluated with Character Error Rate (CER) in each dataset. The dataset details are described in the following subsections.

### 4.1 Kuzushiji dataset

Kuzushiji is a dataset of the pre-modern Japanese in cursive writing style. It is collected and created by the National Institute of Japanese Literature (NIJL). The Kuzushiji_v1 line dataset is a collection of text line images from the first version of the Kuzushiji dataset. The line dataset consists of 25,874 text line images from 2,222 pages of 15 pre-modern Japanese books. There are a total of 4,818 classes. To compare with the state-of-the-art result, we apply the same dataset separation in [24] as shown in Table 1. The testing set is collected from the $15^{th}$ book. The text line images of the $1^{st}$ to $14^{th}$ books are divided randomly into training and validation sets with a ratio of 9:1.

**Table 1.** The detail of the Kuzushiji_v1 line dataset.

|  | **Training set** | **Validation set** | **Testing set** |
|---|---|---|---|
| Text line images | 19,797 | 2,200 | 3,878 |
| Books | $1^{st} \sim 14^{th}$ | | $15^{th}$ |

### 4.2 SCUT-EPT Chinese examination dataset

SCUT-EPT is a new and large-scale offline handwriting dataset extracted from Chinese examination papers [2]. The dataset contains 50,000 text line images from the examination of 2,986 writers, as shown in Table 2. They are randomly divided into 40,000 samples in the training set and 10,000 ones in the test set. We also split the training set into 36,000 samples for the sub-training set and 4,000 for the validation set. There are a total of 4,250 classes, but the number of classes in the training set is just 4,058. Therefore, some classes in the testing set do not appear in the training set. In [2], augmentation from other Chinese character datasets: CASIA-HWDB1.0-1.2 was applied. While the CRNN model covers the entire 7,356 classes, we use only the SCUT-EPT dataset for the training and testing processes.

The character set size of the SCUT-EPT dataset is smaller than that of the Kuzushiji_v1 dataset. However, the SCUT-EPT dataset has some additional problems such as character erasure, text line supplement, character/phrase switching, and noised background.

**Table 2.** The detail of the SCUT-EPT dataset.

|  | **Training set** | **Validation set** | **Testing set** |
|---|---|---|---|
| Text lines images | 36,000 | 4,000 | 10,000 |
| Classes | 4,058 | 4,058 | 3,236 |
| Writers | | 2,986 | |

## 5 Experiments

The configuration of each experiment is declared in Section 5.1. We compare different network configurations on the number of LSTM (BLSTM) layers belonging to the visual extractor and linguistic networks. These configurations are compared with the state-of-the-art methods for the Kuzushiji dataset in Section 5.2 and the SCUT-EPT dataset in Section 5.3. Furthermore, the correctly and wrongly recognized samples are illustrated in Section 5.4.

### 5.1 RNN-Transducer configurations

We use ResNet32 pre-trained on the ImageNet dataset for the visual feature extractor, which allows our model to have good initialization for training. In detail, we use all layers of ResNet32 except the final fully connected layer and replace the final max-pooling layer of ResNet32 with our own max-pooling layer.

With the reading direction from top to bottom in the Kuzushiji Japanese dataset, the width of feature maps is normalized to 1 pixel, and the height is kept unchanged. Meanwhile, for the SCUT-EPT Chinese dataset, which has horizontal text lines, the height of feature maps is normalized to 1 pixel and their width is kept unchanged. At the recurrent layers, we adopt a number of BLSTM layers with 512 hidden nodes. A projection is applied to the sequence by a fully connected layer. Each experiment has a different number of layers and output sizes of the projection. The architecture of our CNN model in the visual extractor is shown in Table 3.

**Table 3.** Network configuration of our CNN model.

| Component | Description |
|---|---|
| Input_1 | $H \times W$ image |
| ResNet32 | All layers except the final max-pooling layer and fully connected layers. |
| MaxPooling_1 | #window: (depend on the dataset) |

An embedding layer is employed at the linguistic context encoder before the recurrent network with the output vector of size 512. The following part is a number of LSTM layers with 512 hidden nodes and a projection layer. These have a similar structure with the recurrent network and the projection layer in the visual extractor network. The joint network projects a concatenated encoded vector from the visual extractor network and linguistic network to 512 dimensions via a fully connected (FC) layer. This concatenated vector is fed to a hyperbolic tangent function and then fed to another FC layer and a softmax layer. The last FC layer has the node size equal to the character set size $Y^*$.

For training the RNN-Transducer model, each training batch and validation batch have 16 and 12 samples. We preprocess and resize the input images to 128-pixel width for the Kuzushiji dataset and 128-pixel height for the SCUT-EPT dataset. We apply augmentations for every iteration, such as random affine transforms and Gaussian noise. Moreover, layer normalization and recurrent variation dropout are employed to prevent over-fitting. We utilize the greedy search in the evaluation.

For updating the RNN-Transducer model parameters, we use an Adam optimizer with an initial learning rate of 2e-4. This learning rate is applied with a linear warmup scheduler in the first epoch and a linear decreasing scheduler in the remaining epochs. The model of each setting is trained with 100 epochs. During training, the best models are chosen according to the best accuracy on the validation set.

To evaluate the effect of the number of LSTM layers and the size of the visual and linguistic encoded vectors, we conduct experiments with 3 different settings. For instance, the number of LSTM layers is from one to three, and the encoded vector sizes are 512 or 1024. All the configurations are described in Table 4.

Table 4. Configurations of RNN-Transducer for SCUT-EPT and Kuzushiji_v1 dataset.

| Name | Configuration |
|---|---|
| S1 | 1 LSTM layer + 512-dimensional encoded vector |
| S2 | 2 LSTM layer + 512-dimensional encoded vector |
| S3 | 3 LSTM layer + 1024-dimensional encoded vector |

### 5.2 Performance on Kuzushiji dataset

Table 4 shows the evaluation of all configurations and the other methods. The lowest CER of 20.33% is from the S1 setting with a single LSTM layer and 512-dimensional encoded vectors. Our best result is lower than the state-of-the-art CER of the AACRN method by 1.15 percentage points. As mentioned before, the RNN-Transducer model is an extension of the CRNN model with the support of a linguistic network. Our model is better than the baseline CRNN model [18] by 8.01 percentage points. Therefore, it proves the efficiency of the linguistic module for the RNN-Transducer model. Furthermore, all other settings of the RNN-Transducer model also have lower CERs than the other previous methods.

Table 5. Character error rates (%) on the test set of the Kuzushiji_v1 dataset.

| Method | CER (%) |
|---|---|
| CRNN [18] | 28.34 |
| Attention-based model [11] | 31.38 |
| AACRN [24] | 21.48 |
| S1 | **20.33** |
| S2 | 20.78 |
| S3 | 21.39 |

### 5.3 Performance on SCUT-EPT dataset

Table 6 shows the results of the RNN-Transducer model in comparison with the other state-of-the-art methods. All our models are better than the other methods without external data and radical knowledge. The lowest CER of 23.15% is from the S3 setting with 3 LSTM layers and 1024 dimensional encoded vectors. As mentioned above, the SCUT-EPT dataset has some additional challenges so that the most complex model-S3 achieves the best performance.

Our best result is better than the state-of-the-art CER of 23.39% by the LODENet method by 0.24 percentage point, while our proposed method does not require prior knowledge of Chinese radicals for the training process. The state-of-the-art CRNN model [2] has a CER of 24.03%, which is slightly inferior to our model. Since the LO-DENet model achieves a CER of 22.39% when text from Wiki corpus and prior radical knowledge are used, our models are expected to be improved further when they are utilized.

**Table 6.** Character error rate (CER) (%) on the test set of the SCUT-EPT dataset.

| Method | CER without external data | CER with external data | Required radical knowledge |
|---|---|---|---|
| CRNN [2] | 24.63 | 24.03 | |
| Attention [2] | 35.22 | 32.96 | |
| Cascaded attention [2] | 51.02 | 44.36 | |
| CTC and MCTC:WE [27] | 23.43 | - | ✓ |
| LODENet [26] | 23.39 | **22.39** | ✓ |
| S1 | 24.34 | - | |
| S2 | 24.44 | - | |
| S3 | **23.15** | - | |

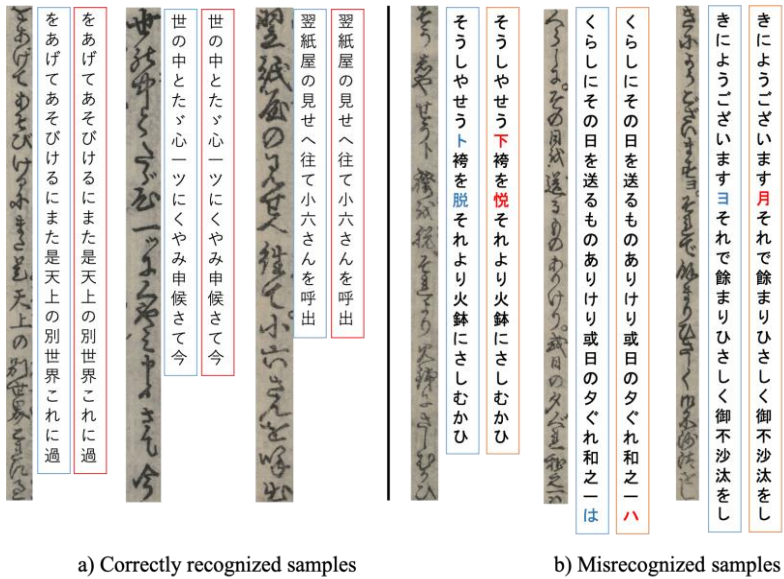

a) Correctly recognized samples    b) Misrecognized samples

**Fig. 2.** Correctly recognized and misrecognized samples in the Kuzushiji_v1 dataset.

### 5.4 Sample visualization

Fig. 2 and Fig. 3 show some correctly recognized and misrecognized samples by the best RNN-Transducer model for the Kuzushiji_v1 and that for SCUT-EPT datasets. For each sample, the blue rectangle below or right shows the ground-truth, and the red one shows the recognition result. As illustrated in Fig. 2b, most of the errors are wrong recognitions of characters in the ground-truth rather than insertion or deletion errors. In Fig. 3a, the results prove that our model can handle challenges in the SCUT-EPT dataset, such as character erasure, text line supplement, and noised background. From the misrecognized samples as shown in Fig. 3b, our model correctly recognizes green-colored characters while the ground-truth text was mislabeled.

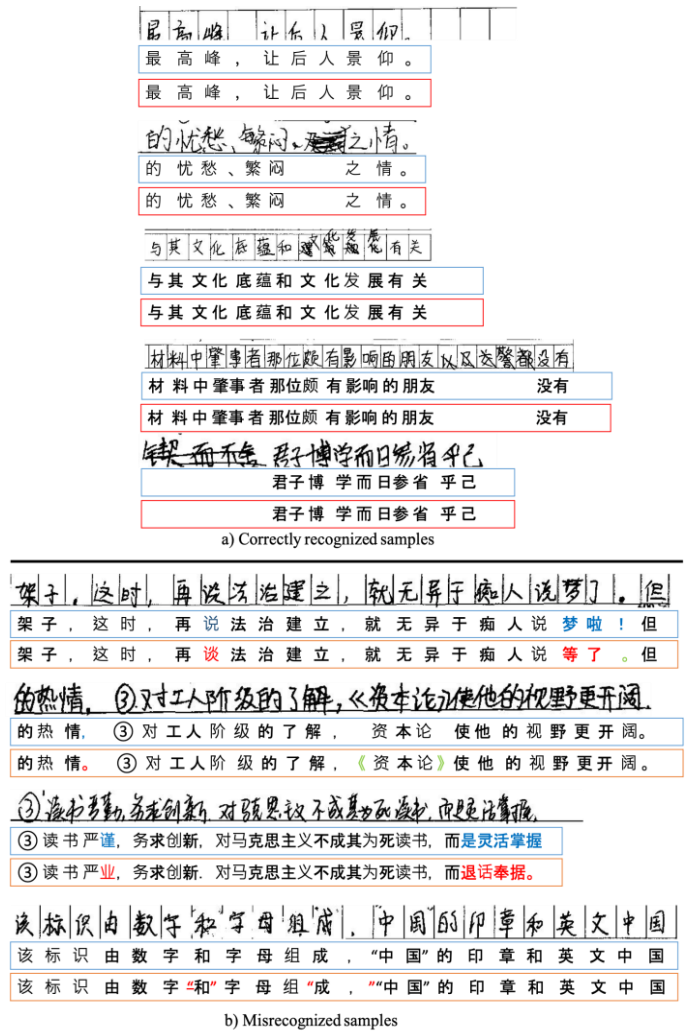

Fig. 3. Correctly recognized and misrecognized samples in the SCUT-EPT dataset.

## 6    Conclusion

In this paper, we have presented an RNN-Transducer model to recognize Japanese and Chinese offline handwritten text line images. The proposed model is effective in modeling both visual features and linguistic context information from the input image. In the experiments, the proposed model archived 20.33% and 23.15% of character error rates on the test set of Kuzushiji and SCUT-EPT datasets, respectively. These results outperform state-of-the-art accuracies on both datasets. In future works, we will conduct experiments of the proposed model with handwritten text datasets in other languages. We also plan to incorporate an external language model into the proposed model.

## Acknowledgement

The authors would like to thank Dr. Cuong Tuan Nguyen for his valuable comments. This research is being partially supported by A-STEP, JST (Grant No: JPMJTM20ML), and the grant-in-aid for scientific research (S) 18H05221 and (A) 18H03597.